\documentclass[runningheads]{llncs}

\usepackage{eccv}

\usepackage{eccvabbrv}

\usepackage{graphicx}
\usepackage{booktabs}
\usepackage{threeparttable}
\usepackage{multirow}
\usepackage{algorithm}
\usepackage{algorithmic}
\usepackage{amsmath}
\usepackage{xcolor}
\usepackage{cite}
\usepackage{pifont}
\usepackage{dsfont}

\newcommand{\INDSTATE}[1][1]{\STATE\hspace{#1\algorithmicindent}}

\usepackage[accsupp]{axessibility}  

\usepackage{hyperref}

\usepackage{orcidlink}

\begin{document}

\title{Diff3DETR: Agent-based Diffusion Model for Semi-supervised 3D Object Detection} 

\titlerunning{Diff3DETR}

\author{Jiacheng Deng\inst{1}\orcidlink{0000-0003-2838-0378} \and
Jiahao Lu\inst{1} \orcidlink{0009-0006-3280-9447} \and
Tianzhu Zhang\inst{1, 2 \dagger ~}\orcidlink{0000-0003-1856-9564}}

\renewcommand{\thefootnote}{\fnsymbol{footnote}}
\footnotetext[4]{Corresponding Author}

\authorrunning{J. Deng et al.}


\institute{
MoE Key Laboratory of Brain-inspired Intelligent Perception and Cognition, University of Science and Technology of China
\\
\and
Deep Space Exploration Lab\\
\email{\tt\small \{dengjc,lujiahao\}@mail.ustc.edu.cn,{tzzhang@ustc.edu.cn}}
}

\maketitle

\begin{abstract}
3D object detection is essential for understanding 3D scenes. Contemporary techniques often require extensive annotated training data, yet obtaining point-wise annotations for point clouds is time-consuming and laborious. Recent developments in semi-supervised methods seek to mitigate this problem by employing a teacher-student framework to generate pseudo-labels for unlabeled point clouds.  However, these pseudo-labels frequently suffer from insufficient diversity and inferior quality. To overcome these hurdles, we introduce an Agent-based Diffusion Model for Semi-supervised 3D Object Detection (Diff3DETR). Specifically, an agent-based object query generator is designed to produce object queries that effectively adapt to dynamic scenes while striking a balance between sampling locations and content embedding. Additionally, a box-aware denoising module utilizes the DDIM denoising process and the long-range attention in the transformer decoder to refine bounding boxes incrementally. Extensive experiments on ScanNet and SUN RGB-D datasets demonstrate that Diff3DETR outperforms state-of-the-art semi-supervised 3D object detection methods. 

\end{abstract}

\section{Introduction}
\label{sec:intro}
3D object detection aims to localize and recognize 3D objects in 3D space to facilitate scene understanding, making it crucial for 3D applications such as autonomous driving~\cite{geiger2012arewereadyfor, arnold2019survey}, AR/VR~\cite{rokhsaritalemi2020review}, and robotic navigation~\cite{ye20173}. The rapid development of deep learning-based methods~\cite{qi2019deep, zhang2020h3dnet, wang2022cagroup3d, liu2021group, misra2021end, he2023hierarchical}, including PointNet~\cite{qi2017pointnet, qi2017pointnet++}, Transformer~\cite{vaswani2017attention}, and DETR~\cite{carion2020end, misra2021end}, has significantly propelled advancements in 3D object detection. However, most existing approaches heavily rely on labeled point cloud data. Manually annotating vast amounts of 3D point cloud scenes is extremely time-consuming and labor-intensive, which could limit the potential for applying 3D object detection in larger-scale scenarios. 
%

To mitigate the dependence on annotated 3D point cloud data, semi-supervised methods~\cite{zhao2020sess,wang20213dioumatch, wang2023not, ho2024diffusion} that utilize a small amount of labeled data alongside a large volume of unlabeled data are gaining attention and rapid development. Methods based on semi-supervised learning~\cite{yang2022survey} leverage the untapped information in unlabeled point clouds to compensate for the information loss due to the scarcity of annotated data, thereby effectively enhancing the detection performance. Existing semi-supervised approaches~\cite{zhao2020sess,wang20213dioumatch, wang2023not, ho2024diffusion} can be broadly categorized into two types: consistency-based methods~\cite{zhao2020sess} and pseudo-label-based methods~\cite{wang20213dioumatch, wang2023not, ho2024diffusion}. For consistency-based methods, the core idea is to encourage consistency in the predictions for data augmented in different ways. For instance, SESS~\cite{zhao2020sess} enforces consensus on object location, semantic category, and size through three consistency losses between model outputs for differently augmented data. However, due to significant noise in model predictions, consistency constraints might lead to suboptimal results.
\begin{figure}[!t]
    \begin{center}
        \includegraphics[width=0.95\textwidth]{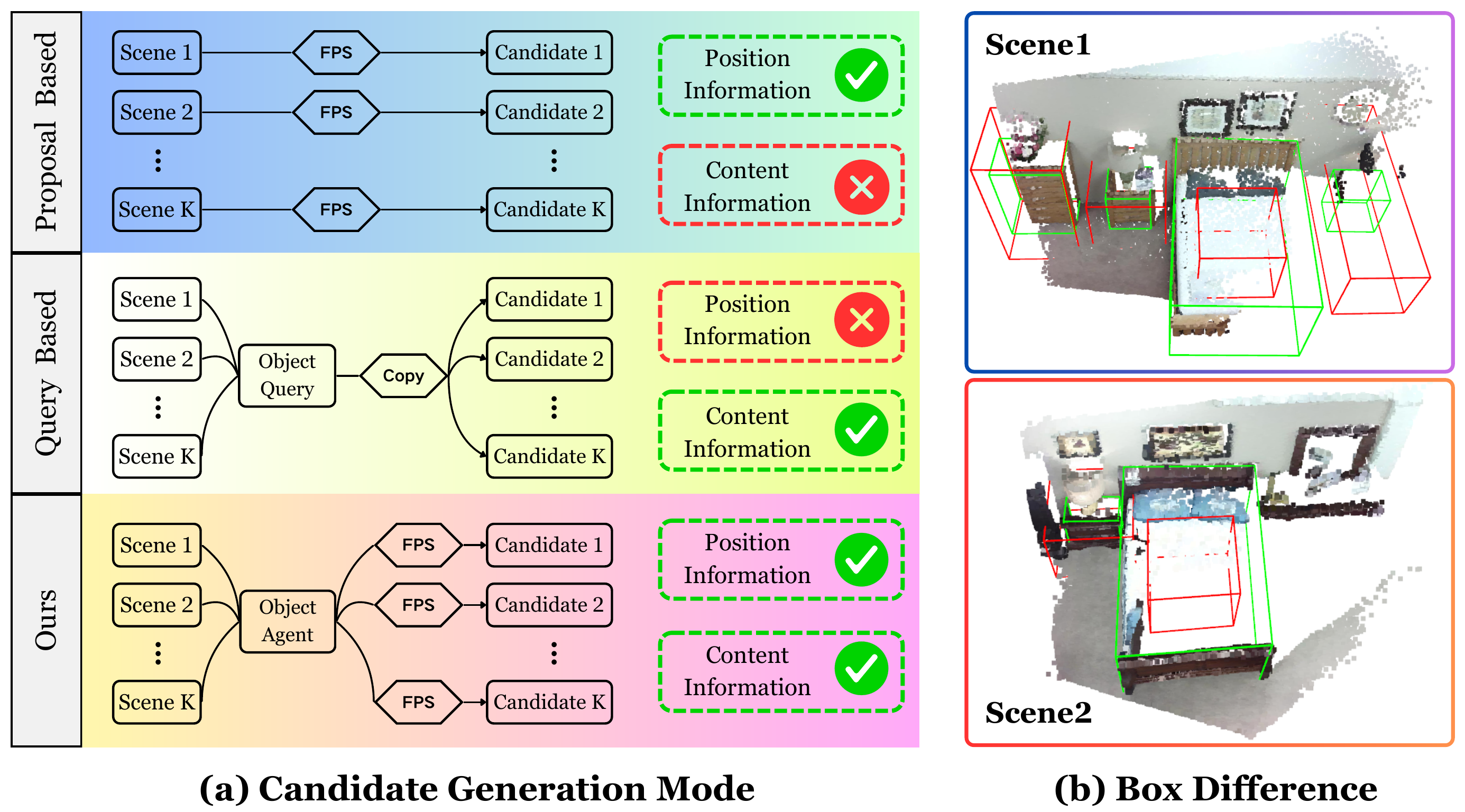}
        \caption{ (a) presents three candidate generation modes: Farthest Point Sampling (FPS), learnable object query, and ours. Our candidate generation mode simultaneously considers the distribution of sampling locations and the learning of content information. (b) displays the geometric differences between initial boxes (in red) and ground truth boxes (in green) in two scenes, highlighting the importance of aggregating features from the correct areas for 3D object detection.}
        \label{teaser}
    \end{center}
\end{figure}

On the other hand, pseudo-label-based methods aim to select high-quality pseudo-labels from the model predictions on unlabeled data and then combine these with labeled data for model training, achieving more accurate detection results. 3DIoUMatch~\cite{wang20213dioumatch} and NESIE~\cite{wang2023not} equip detectors with global and multi-side localization quality estimation modules to assist in pseudo-label filtering and suppression strategies. Diffusion-SS3D~\cite{ho2024diffusion} leverages a diffusion model~\cite{ho2020denoising, song2020denoising, rombach2022high} to randomly initialize noise sizes and labels to enhance the diversity and quantity of pseudo-labels and denoise the noisy boxes to improve the quality of pseudo-labels. The diffusion-based method~\cite{ho2024diffusion} has achieved the best results, marking the diffusion model as a significant trend for the future development of semi-supervised 3D object detection.

Compared to 2D images, the expansive space and sparse distribution of point clouds in 3D scenes result in approximately 90\% of spatial areas lacking point cloud coverage. The existing diffusion-based method~\cite{ho2024diffusion} employs Farthest Point Sampling (FPS) on the point cloud to obtain a fixed number of object candidates,  thus avoiding sampling from empty areas which could lower the recall rate. However, compared to object queries in DETR~\cite{carion2020end}, these candidates struggle to learn sufficient content embedding across scenes. Moreover, the existing diffusion-based method~\cite{ho2024diffusion} aggregates features within the initial boxes. Thus, noisy initial boxes can adversely affect feature aggregation, leading to sub-optimal outcomes.

Based on the discussion above, we identify two critical aspects that need consideration and improvement for building more accurate diffusion-based semi-supervised 3D object detection models:
1) \textit{How to effectively model object candidates?}
As shown in Figure~\ref{teaser}(a), existing 3D object detection methods~\cite{qi2019deep, xie2020mlcvnet, misra2021end,liu2021group, shen2023v} primarily obtain candidates through two approaches: Farthest Point Sampling (FPS) and learnable object query~\cite{carion2020end}. FPS sampling is more likely to distribute candidates across areas where objects are located, reducing the rate of empty sampling. However, it cannot learn content embedding across scenes to aggregate object features effectively. On the other hand, learnable object queries can update and learn across scenes in the dataset but have a higher rate of sampling empty spaces, which lowers the object recall rate. Therefore, balancing the sampling position and the content learning is critical to modeling object candidates effectively.
2) \textit{How to aggregate the correct features to assist in the incremental refinement of noisy initial boxes?}
Existing diffusion-based methods~\cite{chen2023diffusiondet, ho2024diffusion} generate initial bounding boxes using random Gaussian noise. However, as illustrated in Figure~\ref{teaser}(b), these noisy initial boxes significantly differ from the ground truth object boxes. Consequently, the aggregated features often contain substantial errors, making it challenging to iteratively predict accurate object boxes. Therefore, it is crucial to appropriately expand the receptive field around the initial boxes to more accurately locate the correct target area and aggregate features, aiding in the denoising and correction of noisy boxes.

To achieve the above goals, we propose an agent-based diffusion model in a unified DETR architecture for semi-supervised 3D object detection, namely \textbf{Diff3DETR}, which consists of an agent-based object query generator and a box-aware denoising module.
Overall, our method employs the mean teacher framework, integrating the diffusion model and DETR architecture within a unified model. The randomness of the diffusion process generates a greater quantity of pseudo-labels. Simultaneously, the agent-based object query generator creates object queries that balance sampling locations and content embedding. In the box-aware denoising module, features are aggregated within a long-range perceptive field to iteratively optimize noisy boxes, achieving accurate object predictions.
More specifically, the agent-based object query generator initially establishes learnable object agents to obtain satisfactory content embeddings, where object agents could dynamically adapt to specific scenes through interaction with scene features. The object queries are derived through linear interpolating FPS-sampled locations with object agents. In the box-aware denoising module, the DDIM~\cite{song2020denoising} denoising process and transformer decoder are ingeniously intertwined. Moreover, to ensure each object query focuses on the designated object area during the query process, we bind object queries with the noisy box locations, enhancing the positional dependency of object queries. 

To sum up, the contributions of this work can be summarized as follows: 
(1) We introduce an agent-based diffusion model within a unified DETR framework, which includes an agent-based object query generator and a box-aware denoising module. To the best of our knowledge, this is the first diffusion-based DETR framework in the semi-supervised 3D object detection field. 
(2) We develop an agent-based object query generator to generate object queries that better adapt to dynamic scenes while balancing sampling locations and content embedding. Additionally, we design a box-aware denoising module that leverages the incremental refinement capabilities of the DDIM denoising process and the long-range attention of the transformer decoder to denoising initial boxes for accurate 3D object detections. 
(3) Extensive experimental results on the ScanNet and SUN RGB-D datasets demonstrate that Diff3DETR achieves superior performance and outperforms existing state-of-the-art methods.

\section{Related Work}
\label{sec:relatedwork}
In this section, we provide a concise overview of methodologies related to diffusion models for perception tasks and semi-supervised 3D object detection.


\textbf{Diffusion Models for Perception Tasks.}
Diffusion models have demonstrated remarkable success in image generation~\cite{ho2020denoising,sohl2015deep,kingma2021variational,cao2024survey, peebles2023scalable}. Consequently, researchers have begun exploring the integration of diffusion models into perception tasks. Pix2Seq-D~\cite{chen2023generalist} utilizes the Bit Diffusion model~\cite{chen2022analog} to conduct panoptic segmentation~\cite{kirillov2019panoptic} tasks across images and videos. ~\cite{mukhopadhyay2023diffusion} utilizes diffusion models, showcasing their efficacy in extracting discriminative features for classification tasks. DiffusionDet~\cite{chen2023diffusiondet} frames object detection as a noise-to-box task, wherein high-quality object bounding boxes are generated by progressively denoising randomly generated proposals. In line with this work, Diffusion-SS3D~\cite{ho2024diffusion} harnesses the power of diffusion models in semi-supervised 3D object detection settings, aiming to offer a novel approach to generate more dependable pseudo-labels. In this work, we extend the approach of Diffusion-SS3D by proposing the first diffusion-based DETR framework, which includes an agent-based object query generator and a box-aware denoising module.

\textbf{Semi-supervised 3D Object Detection.}
While fully supervised methods~\cite{qi2019deep,xie2020mlcvnet,gwak2020generative,liu2021group,misra2021end, wang2023long} have demonstrated superior performance, they are often constrained by the labor-intensive process of bounding box annotations, presenting significant practical limitations. Consequently, several semi-supervised 3D object detection techniques~\cite{zhao2020sess,wang20213dioumatch,ho2024diffusion, liu2023hierarchical, wu2024semi, griffiths2020finding} have emerged to address this issue. SESS~\cite{zhao2020sess} represents the pioneering effort in semi-supervised 3D object detection. By enforcing consistency between the outputs of the mean teacher~\cite{tarvainen2017mean}, SESS effectively learns from unlabeled data. 3DIoUMatch~\cite{wang20213dioumatch} introduces confidence-based filtering and IoU prediction strategies to select high-quality pseudo-labels generated by the teacher model. However, many existing methods heavily rely on the teacher model for pseudo-label generation, limiting their ability to identify bounding boxes beyond the scope of teacher model predictions. To mitigate this challenge, models based on the Diffusion model have been proposed. Diffusion-SS3D~\cite{wang20213dioumatch} leverages the diffusion model for semi-supervised 3D object detection, treating the task as a denoising process to enhance the quality of pseudo-labels. In this work, we adhere to the diffusion-based methodology and employ the agent-based object query generator to produce object queries better suited for complex and dynamic environments. Additionally, we devise the box-aware denoising module to enhance refinement capabilities.

\section{Methodology}
\label{sec:method}
In this section, we introduce the details of our Diff3DETR. Section~\ref{pre} covers the preliminaries. Section~\ref{sec_diff3detr} describes the Diff3DETR framework, focusing on how the teacher detector generates high-quality pseudo-labels to assist the student detector's training and the computational specifics of both models. Section~\ref{sec_detector} elaborates on the structural details of the Diff3DETR detector.

\subsection{Preliminaries}
\label{pre}
\textbf{Semi-supervised 3D object detection.}
Given the point cloud of a scene as input, the goal of 3D object detection is to classify and localize amodal 3D bounding boxes for objects within the point cloud. The point cloud data is $\mathbf{x} \in \mathbb{R}^{n \times 3}$, where $n$ denotes the number of points. Within the semi-supervised framework, we are provided with $N$ training samples. The samples involve a set of $N_l$ labeled scenes $\{\mathbf{x}_i^l, \mathbf{y}_i^l\}_{i=1}^{N_l}$ and a larger set of $N_u$ unlabeled scenes $\{\mathbf{x}_i^u\}_{i=1}^{N_u}$. The ground-truth annotations $\mathbf{y}_i^l$ encompass $K$ objects $\{\mathbf{b}_k^l, l_k\}_{k=1}^{K}$ within $\mathbf{x}_i^l$, with $\mathbf{b}$ and $l$ signifying a collection of bounding box parameters and semantic class labels with a total of $N_{cls}$ classes. Specifically, the bounding box $\mathbf{b}$ is formulated as $\mathbf{b} = \{\mathbf{b}_c, \mathbf{b}_s, \mathbf{b}_o\}$, where $\mathbf{b}_c = \{c_x, c_y, c_z\}$ delineates the centroid coordinates, $\mathbf{b}_s = \{s_l, s_w, s_h\}$ represents the object dimensions, and $\mathbf{b}_o$ is the object orientation along the upright-axis. 

\textbf{Diffusion model.}
Diffusion models, a class inspired by non-equilibrium thermodynamics~\cite{de2013non}, present a novel approach in generating data by progressively introducing noise into the data samples. This process is mathematically modeled as a Markov chain~\cite{norris1998markov} consisting of $T$ diffusion steps, with the forward diffusion process described by:
\begin{equation}
q(\mathbf{z}_T|\mathbf{z}_0) = \mathcal{N}(\mathbf{z}_T| \sqrt{\bar{\alpha}_T}\mathbf{z}_0, (1 - \bar{\alpha}_T)\mathbf{I}),
\label{denoising}
\end{equation}
where \( \mathbf{I} \) is identity matrix. In this equation, the original data sample $\mathbf{z}_0$ is transformed into a noisy latent representation $\mathbf{z}_T$ through the application of additive noise. The noise scale \( \alpha_s := 1 - \beta_s \) and \( \bar{\alpha}_T := \prod_{s=1}^{T}\alpha_s \) are computed as the product of individual noise scales from the first diffusion step to step $T$, and $\beta_s$ is a predetermined variance schedule. A neural network $f_\theta(\mathbf{z}_T)$ is then trained to reverse this diffusion process by predicting the noiseless data $\mathbf{z}_0$ from the noisy data $\mathbf{z}_T$, optimizing an $L_2$ loss objective:
\begin{equation}
\mathcal{L}_{train} = \frac{1}{2}\left\|f_\theta(\mathbf{z}_T) - \mathbf{z}_0\right\|^2
\label{ddim_loss}
\end{equation}
During the inference phase, the model iteratively reconstructs the original data sample from its noisy counterpart by iteratively applying an update rule. The sequence of transformations $ \mathbf{z}_T \rightarrow \mathbf{z}_{T-\Delta} \rightarrow \ldots \rightarrow \mathbf{z}_0 $ refines the data sample at each step until the original data is retrieved. A detailed formulation of diffusion models is provided in the supplementary materials. 

\begin{figure*}[t]
  \centering
   \includegraphics[width=0.75\textwidth]{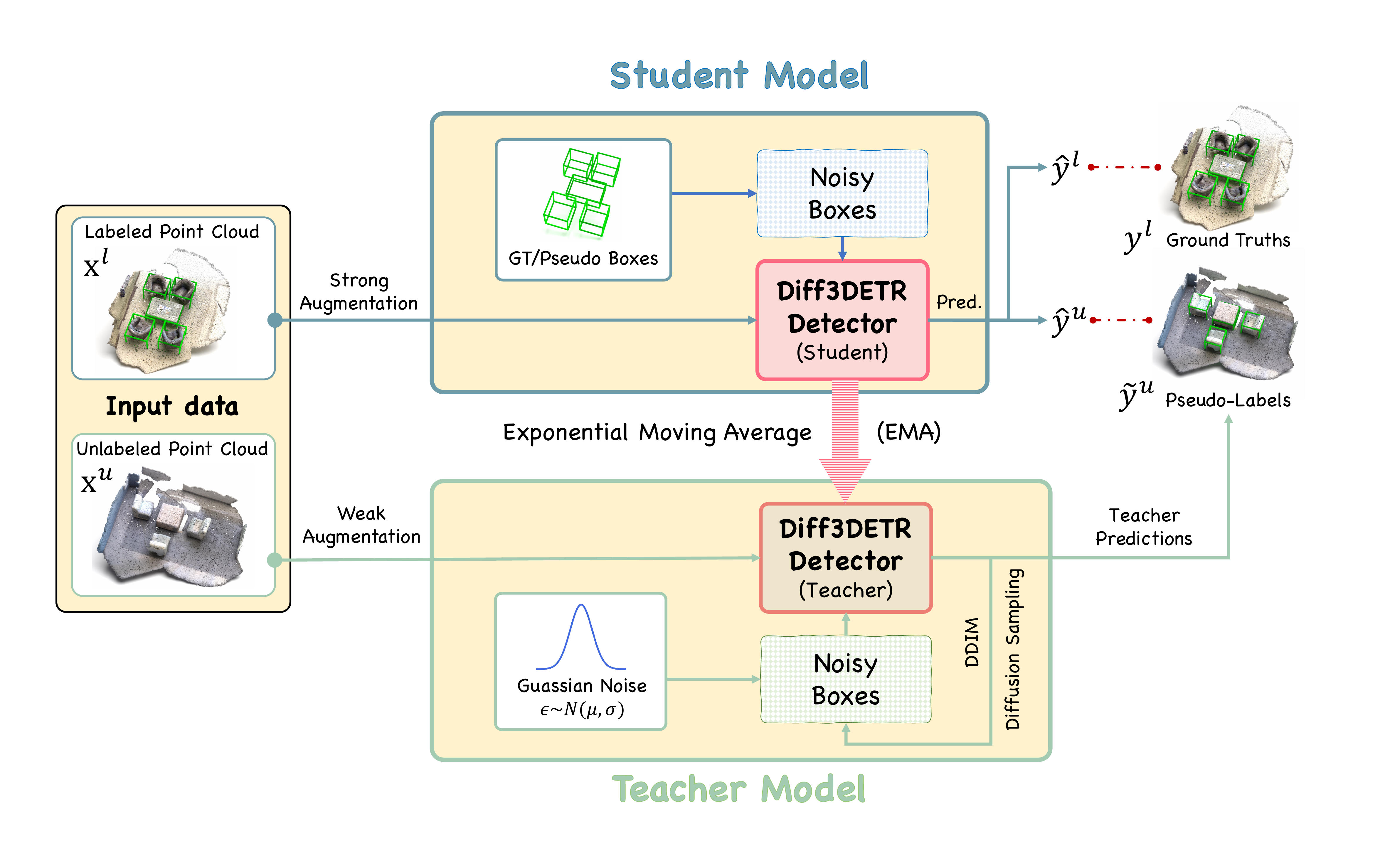}
   \caption{\textbf{The framework of Diff3DETR.} Diff3DETR adopts the framework of the mean teacher~\cite{tarvainen2017mean}, consisting of a student model and a teacher model. The student and teacher models start from GT/pseudo boxes and Gaussian noise, respectively, gradually adding noise to generate noisy boxes and ultimately predicting the accurate object boxes through the Diff3DETR detector. The student model updates its parameters under the supervision of ground truths and pseudo-labels, while the teacher model updates its parameters using an Exponential Moving Average (EMA) strategy. }
   \label{fig:pipeline}
\end{figure*}

\subsection{The Teacher-Student Framework of Diff3DETR}
\label{sec_diff3detr}

The proposed Diff3DETR is a novel diffusion-based framework for semi-supervised 3D object detection. As illustrated in Figure~\ref{fig:pipeline}, Diff3DETR first processes the input point cloud through data augmentation of varying intensities, subsequently feeding them into two branches: the student model and the teacher model. The teacher model introduces Gaussian noise to the 3D bounding boxes and reverses the diffusion process as in DDIM~\cite{song2020denoising} to generate reliable pseudo bounding boxes. 
For the student model, the student model takes both labeled and unlabeled point clouds as input and adds Gaussian noise to the ground truth boxes and pseudo boxes to obtain noisy boxes. The student Diff3DETR detector then takes these noisy boxes and the scene point cloud as input, directly predicting the final object boxes without undergoing a multi-round diffusion sampling process. The prediction results $\hat{\mathbf{y}}^l$ and $\hat{\mathbf{y}}^u$ are supervised by ground truths $\mathbf{y}^l$ and pseudo-labels $\tilde{\mathbf{y}}^u$, respectively. Furthermore, the parameters of the teacher Diff3DETR detector are updated using an Exponential Moving Average (EMA) strategy based on the parameters of the student Diff3DETR detector.

To more clearly describe the computational processes of the teacher model and the student model, we detail the computation specifics for both in Algorithm~\ref{alg1} and Algorithm~\ref{alg2}, respectively. Compared to the teacher model, the student model mainly differs in the following aspects: 1) the generation of the noisy box's size and label is achieved by incrementally adding Gaussian noise to the GT/pseudo boxes; 2) instead of using a multi-step DDIM diffusion sampling process for denoising the noisy boxes, it directly predicts accurate boxes and computes the loss with ground truths and pseudo-labels. The overall loss function \( \mathcal{L} \) for the student model is defined as follows:
\begin{equation}
    \mathcal{L} = \mathcal{L}_l(x^l, y^l) + \lambda\mathcal{L}_u(x^u, \tilde{y}^u),
\end{equation}
where \( \mathcal{L}_l \) and \( \mathcal{L}_u \) are the detection loss functions for labeled samples and unlabeled samples, respectively. Both loss functions are used in 3DIoUMatch~\cite{wang20213dioumatch} for bounding box regression and classification, combined through a loss weight hyper-parameter \( \lambda \). Model inference is completed through the student model by adding DDIM diffusion sampling for multi-step iterative denoising, similar to the inference step in Algorithm~\ref{alg1}.

\begin{figure*}[t]
  \centering
   \includegraphics[width=0.9\linewidth]{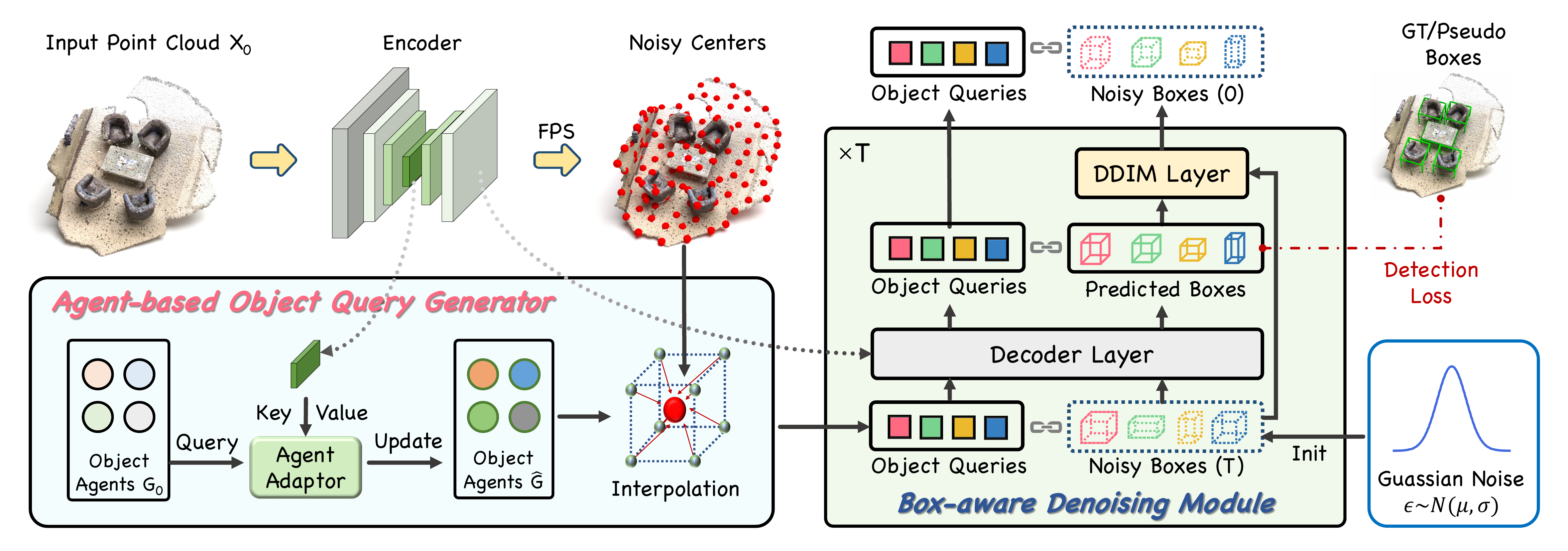}
   \caption{\textbf{The overall architecture of the Diff3DETR detector.} 
   Input point clouds undergo downsampling and feature extraction, with FPS selecting noisy centers. The agent-based object query generator sets learnable agents interacting with scene features and generates object queries through trilinear interpolation with these centers. Concurrently, noisy boxes initialized with Gaussian noise and object queries are processed in the box-aware denoising module. This module updates queries and predicts boxes, aided by the DDIM layer for iterative denoising. }
   \label{fig:detector}
\end{figure*}

\subsection{The Detector of Diff3DETR}
\label{sec_detector}
The overall architecture of the Diff3DETR detector is depicted in Figure~\ref{fig:detector}. Initially, input point clouds are downsampled and feature-extracted using a PointNet++ encoder~\cite{qi2017pointnet}. These downsampled point clouds are further processed with Farthest Point Sampling (FPS) to identify noisy centers, which serve as preliminary center points for object boxes. The coordinates of these noisy centers are then interpolated in an agent-based object query generator to form object queries for detecting scene targets. Noisy boxes are comprised of noisy centers, along with noisy sizes and noisy labels, which are directly initialized using Gaussian noise. Object queries and noisy boxes undergo prediction of the actual boxes and gradual iterative denoising of noisy boxes within a box-aware denoising module through a decoder layer and the DDIM~\cite{song2020denoising} layer, respectively.

\textbf{Agent-based object query generator.}
The agent-based object query generator uniformly samples the normalized scene space at a resolution of \(L \times W \times H\) grid points and assigns a learnable variable to each grid point. The object agents are formed by adding each grid point's learnable variable to its grid point position coordinates encoded by a two-layer MLP. The initial object agents are denoted as \(G_0 \in \mathbb{R}^{N_G \times C}\), where $N_G = L\times W \times H$ denotes the number of object agents and $C$ represents the number of feature channels.

\begin{figure*}[!b]
  \centering
   \includegraphics[width=0.8\linewidth]{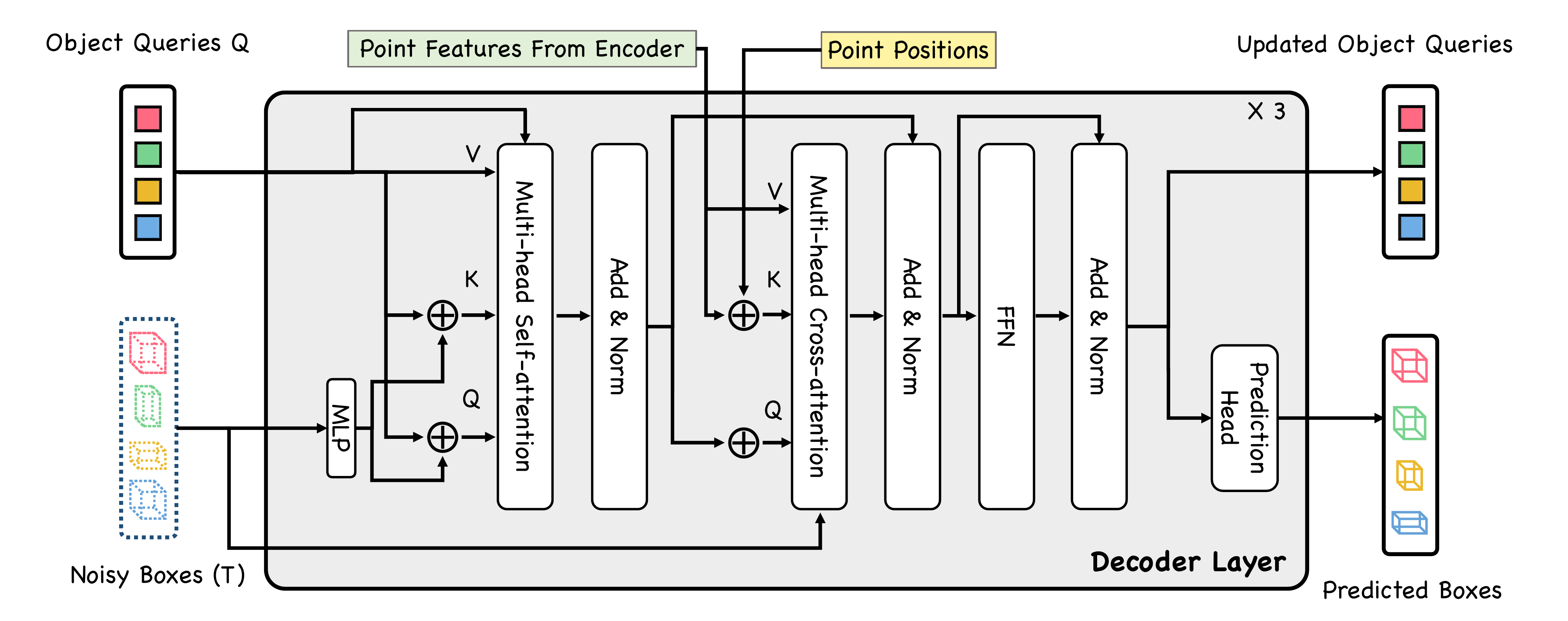}
   \caption{\textbf{The architecture of the decoder layer.} }
   \label{fig:decoder}
\end{figure*}

 To facilitate dynamic adaptability of object agents across various scenes, we design an agent adaptor based on the attention mechanism~\cite{vaswani2017attention}, which can be formulated as follows:
\begin{equation}
    \label{agent adaptor}
    Q = \text{Attention}(Q, K, V) = \text{softmax}\left(\frac{QK^T}{\sqrt{d_k}}\right)V,
\end{equation}
where \(Q\), \(K\), and \(V\) represent the queries, keys, and values, respectively, and \(d_k\) denotes the dimension of the keys. In this context, queries represent $G_0$, and keys and values correspond to the high-level point cloud semantic features from the encoder's intermediate layers. The updated object agents, denoted as \(\hat{G}\), are then interpolated through the noisy centers' coordinates $\mathbf{b}^{noise}_c$ to obtain the object queries \(\hat{O} \in \mathbb{R}^{N_Q \times C}\), where $N_Q$ represents the number of object queries and $C$ is the number of feature channels.

\begin{figure}[!t]
  \centering
  \begin{minipage}{.48\textwidth}
    \begin{algorithm}[H]
      \fontsize{6.5pt}{8.5pt}\selectfont 
      \caption{Teacher Model}
      \label{alg1}
      \begin{algorithmic}
        \STATE \textbf{Input:} point cloud, steps, T  
        \STATE \textbf{Output:} pseudo labels
        
        \STATE  \textcolor[rgb]{0.22,0.7,0.18}{*Extract points and features}
        \STATE pts, feats = Teacher.encoder(pc)
        \STATE  \textcolor[rgb]{0.22,0.7,0.18}{*Update object agents with the input scene }
        \STATE object\_agents = \textcolor[rgb]{0.85,0.185,0.52}{update}(object\_agents, feats)

        \STATE
        \STATE \textcolor[rgb]{0.22,0.7,0.18}{*Generate object centers}
        \STATE centers = \textcolor[rgb]{0.85,0.185,0.52}{FPS}(pts)

        \STATE \textcolor[rgb]{0.22,0.7,0.18}{*Object query interpolation}
        \STATE queries = \textcolor[rgb]{0.85,0.185,0.52}{interpolate}(object\_agents, centers)

        \STATE
        \STATE \textcolor[rgb]{0.22,0.7,0.18}{*Noisy sizes and class labels}
        \STATE sizes\_t = normal(mean=0, std=1)
        \STATE labels\_t = normal(mean=0, std=1)
        
        \STATE
        \STATE \textcolor[rgb]{0.22,0.7,0.18}{*Uniform sample DDIM times}
        \STATE times = \textcolor[rgb]{0.85,0.185,0.52}{reversed}(linespace(-1, T, steps))
        \STATE time\_pairs = \textcolor[rgb]{0.85,0.185,0.52}{list}(\textcolor[rgb]{0.85,0.185,0.52}{zip}(times[:-1], times[1:])) 
        \STATE
        \STATE for t\_cur, t\_next in zip(time\_pairs):
        \INDSTATE \textcolor[rgb]{0.22,0.7,0.18}{*Generate noisy boxes}
        \INDSTATE boxes\_t = \textcolor[rgb]{0.85,0.185,0.52}{random\_match}((queries, centers), \INDSTATE \quad \quad \quad \quad (sizes\_t, labels\_t))
        
        \INDSTATE \textcolor[rgb]{0.22,0.7,0.18}{*Predict pseudo labels}
        \INDSTATE pls = Teacher.decoder(queries, feats,
        \INDSTATE \quad \quad \quad \quad boxes\_t, t\_cur)
        
        \INDSTATE \textcolor[rgb]{0.22,0.7,0.18}{*Estimate boxes\_t at t\_next}
        \INDSTATE boxes\_t = \textcolor[rgb]{0.85,0.185,0.52}{ddim}(boxes\_t, pls, t\_cur, t\_next)

        \INDSTATE \textcolor[rgb]{0.22,0.7,0.18}{*Box renewal}
        \INDSTATE boxes\_t = \textcolor[rgb]{0.85,0.185,0.52}{box\_renewal}(boxes\_t)
        \STATE
        
        \STATE \textcolor[rgb]{0.22,0.7,0.18}{*Filtering}
        \STATE pls = \textcolor[rgb]{0.85,0.185,0.52}{filter}(pls)

        \STATE \textbf{Return} pls
      \end{algorithmic}
    \end{algorithm}
  \end{minipage}%
  \hfill
  \begin{minipage}{.48\textwidth}
    \begin{algorithm}[H]
      \fontsize{6.5pt}{8.5pt}\selectfont 
      \caption{Student Model}
      \label{alg2}
      \begin{algorithmic}
        \STATE \textbf{Input:} point cloud, gts, pls 
        \STATE \textbf{Output:} predictions
        
        \STATE  \textcolor[rgb]{0.22,0.7,0.18}{*Extract points and features}
        \STATE pts, feats = Student.encoder(pc)
        \STATE  \textcolor[rgb]{0.22,0.7,0.18}{*Update object agents with the input scene }
        \STATE object\_agents = \textcolor[rgb]{0.85,0.185,0.52}{update}(object\_agents, feats)
        \STATE \textcolor[rgb]{0.22,0.7,0.18}{*Pad bounding boxes}
        \STATE sizes = \textcolor[rgb]{0.85,0.185,0.52}{prepare\_size}(gts, pls)
        \STATE labels = \textcolor[rgb]{0.85,0.185,0.52}{prepare\_label}(gts, pls)
        \STATE \textcolor[rgb]{0.22,0.7,0.18}{*Signal scaling}
        \STATE sizes = (sizes*2 -1)* size\_scale
        \STATE labels = (labels*2 -1)* label\_scale 
        
        \STATE \textcolor[rgb]{0.22,0.7,0.18}{*Generate object centers}
        \STATE centers = \textcolor[rgb]{0.85,0.185,0.52}{FPS}(pts)

        \STATE \textcolor[rgb]{0.22,0.7,0.18}{*Object query interpolation}
        \STATE queries = \textcolor[rgb]{0.85,0.185,0.52}{interpolate}(object\_agents, centers)

        \STATE \textcolor[rgb]{0.22,0.7,0.18}{*Corrupt GT bounding boxes}
        \STATE t = randint(0, T)
        \STATE eps = normal(mean=0, std=1)
        \STATE sizes\_crpt = sqrt(alpha\_cumprod(t))*sizes
        \INDSTATE \quad \quad +sqrt(1-alpha\_cumprod(t))*eps
        \STATE labels\_crpt = sqrt(alpha\_cumprod(t))*labels
        \INDSTATE \quad \quad +sqrt(1-alpha\_cumprod(t))*eps
        \STATE \textcolor[rgb]{0.22,0.7,0.18}{*Generate noisy boxes}
        \STATE boxes\_crpt = \textcolor[rgb]{0.85,0.185,0.52}{match}((queries, centers), \INDSTATE \quad \quad \quad \quad (sizes\_crpt, labels\_crpt))
        \STATE \textcolor[rgb]{0.22,0.7,0.18}{*Predict object candidates}
        \STATE preds = Student.decoder(queries, feats, \INDSTATE \quad \quad \quad \quad boxes\_crpt, t)
        \STATE \textcolor[rgb]{0.22,0.7,0.18}{*Update student by (pseudo) ground truths}
        \STATE loss = \textcolor[rgb]{0.85,0.185,0.52}{detector\_loss}(preds, gts, pls)
        \STATE Student = \textcolor[rgb]{0.85,0.185,0.52}{grad\_update}(Student, loss)
        \STATE \textcolor[rgb]{0.22,0.7,0.18}{*Update teacher via exponential moving average}
        \STATE Teacher = \textcolor[rgb]{0.85,0.185,0.52}{ema\_update}(Teacher, Student)
        \STATE \textbf{Return} preds
      \end{algorithmic}
    \end{algorithm}
  \end{minipage}
\end{figure}

\textbf{Box-aware denoising module.}
Noisy boxes comprise three components: noisy centers, noisy sizes, and noisy semantic labels. As shown in Figure~\ref{fig:decoder}, object queries and noisy boxes are fed into the decoder layer, where object queries are updated through self-attention and cross-attention mechanisms to predict accurate boxes. 
Given object query set $\hat{O} = \{q_i\}$, noisy boxes $\mathbf{b^{noise}} = \{\mathbf{b}_c^{noise}, \mathbf{b}_s^{noise}, \mathbf{b}_o^{noise}\}$ and point cloud features $P = \{p_k\}$, the output feature of the multi-head self-attention of each query element is the aggregation of the values that weighted by the attention weights, formulated as:
\begin{equation}
\text{Self-Att}(q_i, \{q_k\}) = \sum_{h=1}^{H} W_h \left( \sum_{k=1}^{K} A_{h,i,k} \cdot V_h q_k \right), 
\label{self-attention}
\end{equation}
\begin{equation}
A_{h,i,k} = \frac{\exp\left( (Q_h (q_i+\mathbf{MLP}(b^{noise}_i)))^\top (U_h (q_k+\mathbf{MLP}(b^{noise}_k))) \right)}{\sum_{k=1}^{K} \exp\left(  (Q_h (q_i+\mathbf{MLP}(b^{noise}_i)))^\top (U_h (q_k+\mathbf{MLP}(b^{noise}_k))) \right)}, 
\label{self-add}
\end{equation}
where $h$ indexes over attention heads, $A_h$ is the attention weight, $Q_h, V_h, U_h, W_h$ indicate the query projection weight, value projection weight, key projection weight, and output projection weight, respectively. The output feature of the multi-head cross-attention of each object query are formulated as:
\begin{equation}
\text{Cross-Att}(q_i, \{p_k\}) = \sum_{h=1}^{H} W_h \left( \sum_{k=1}^{K} \bar{A}_{h,i,k} \cdot V_h p_k \right), 
\label{cross-attention}
\end{equation}
\begin{equation}
\bar{A}_{h,i,k} = \frac{\exp\left( (Q_h q_i)^\top (U_h p_k)+\mathbf{R}_{i,k} \right)}{\sum_{k=1}^{K} \exp\left( (Q_h q_i)^\top (U_h p_k)+\mathbf{R}_{i,k} \right)},
\label{cross-add}
\end{equation}
where \(\mathbf{R}_{i,k}\) represents the 3D Vertex Relative Position Encoding (3DV-RPE) of the \(i\)th object query's corresponding noisy box with respect to the \(k\)th point in the point cloud. Inspired by the V-DETR~\cite{shen2023v} approach, 3DV-RPE encodes a point by its relative position to the target object and is crucial for augmenting the transformers to capture the spatial context of the tokens. For further details, please refer to V-DETR~\cite{shen2023v}. 
The denoising process for noisy boxes occurs in the DDIM layer as detailed in Algorithm~\ref{alg1}, predicting the noisy boxes for next step \(t_{next}\) based on the predicted boxes and noisy boxes at current step \(t_{cur}\).


\section{Experiments}
\label{sec:Experiments}
\subsection{Datasets.}
In our study, we conduct evaluations on two primary datasets: ScanNet~\cite{dai2017scannet} and SUN RGB-D~\cite{song2015sun}, employing evaluation protocols from existing semi-supervised 3D object detection literature. ScanNet is an established dataset for 3D indoor scene benchmarking and is composed of 1,201 training and 312 validation scenes, reconstructed from 2.5 million high-resolution RGB-D images. Our study places an emphasis on the 18 semantic classes as aligned with prior studies. The SUN RGB-D dataset, another significant 3D benchmark, consists of 5,285 training scenes and 5,050 validation scenes. We assess the models across 10 object classes.

\subsection{Evaluation metrics.}
For the model evaluation, we split the datasets into partitions with various proportions of labeled and unlabeled data to support semi-supervised learning (SSL). Specifically, we allocate 5\%, 10\%, 20\%, and 100\% of labeled data for the ScanNet evaluation, and 1\%, 5\%, 10\%, and 20\% for SUN RGB-D. Performance metrics are calculated using the mean Average Precision (mAP). mAP@0.25 at an Intersection over Union (IoU) threshold of 0.25 and mAP@0.5 are reported. These metrics provide insights into precision at a granular level for object detection tasks. The evaluation is conducted across three random data splits to ensure robustness, and report averaged performance and the standard deviation.

\subsection{Implementation details}
For the detector, we establish a grid of object agents with \( (L, W, H) = (16, 16, 4) \) and set the number of noisy centers and object queries to 128. 
For the diffusion process, we set the maximum timesteps to 1000. Like Diffusion-SS3D~\cite{ho2024diffusion}, we set the mean of the sampling sizes to 0.25. We set the random sampling mean for noisy labels to the inverse of the respective class number. Our teacher model employs a dual-step DDIM sampling technique \( (T=2) \) to generate pseudo-labels and produce final evaluation results. 

\begin{table}[!htbp]
  \begin{center}
    \footnotesize
    \setlength\tabcolsep{2pt}
\caption{\textbf{Results on ScanNet val dataset under different ratios of labeled data.} The best is denoted by \textbf{boldface}, while the second best is \underline{underlined}.} 
\label{table:ScanNet}
\scalebox{0.75}{
\begin{tabular}{c|c|c|c|c|c|c|c|c}
\toprule
\multirow{2}*{Model} & \multicolumn{2}{c|}{5\%} & \multicolumn{2}{c|}{10\%} & \multicolumn{2}{c|}{20\%} & \multicolumn{2}{c}{100\%}  \\
\cline{2-3}\cline{4-5} \cline{6-7} \cline{8-9} 
 & mAP@0.25 & mAP@0.5 & mAP@0.25 & mAP@0.5 & mAP@0.25 & mAP@0.5 & mAP@0.25 & mAP@0.5 \\
\midrule
VoteNet\cite{qi2019deep} & 27.9\(\pm\)0.5 & 10.8\(\pm\)0.6 & 36.9\(\pm\)1.6 & 18.2\(\pm\)1.0 & 46.9\(\pm\)1.9 & 27.5\(\pm\)1.2  & 57.8 & 36.0 \\
SESS\cite{zhao2020sess} & 32.0\(\pm\)0.7 & 14.4\(\pm\)0.7 & 39.5\(\pm\)1.8 & 19.8\(\pm\)1.3 & 49.6\(\pm\)1.1 & 29.0\(\pm\)1.0  & 61.3 & 39.0 \\
3DIoUMatch\cite{wang20213dioumatch} & 40.0\(\pm\)0.9 & 22.5\(\pm\)0.5 & 47.2\(\pm\)0.4 & 28.3\(\pm\)1.5 & 52.8\(\pm\)1.2 & 35.2\(\pm\)1.1  & 62.9 & 42.1 \\
NESIE\cite{wang2023not} & 40.5\(\pm\)1.1 & 23.8\(\pm\)0.8 & 48.8\(\pm\)0.9 & 31.1\(\pm\)1.1 & 54.5\(\pm\)0.8 & 37.3\(\pm\)0.5  & 63.8 & 44.1 \\
Diffusion-SS3D\cite{ho2024diffusion} & \underline{43.5\(\pm\)0.2} & \underline{27.9\(\pm\)0.3} & \underline{50.3\(\pm\)1.4} & \underline{33.1\(\pm\)1.5} & \underline{55.6\(\pm\)1.7} & \underline{36.9\(\pm\)1.4} & \underline{64.1} & \underline{43.2} \\
\textbf{Ours} & \textbf{45.1\(\pm\)0.5} & \textbf{29.2\(\pm\)0.5} & \textbf{51.6\(\pm\)1.2} & \textbf{34.2\(\pm\)0.9} & \textbf{57.0\(\pm\)1.5} & \textbf{38.2\(\pm\)0.7} & \textbf{65.7} & \textbf{44.9} \\
\bottomrule
\end{tabular}
}
\end{center}
\end{table}

\begin{table}[!htbp]
  \begin{center}
    \footnotesize
    \setlength\tabcolsep{2pt}
\caption{\textbf{Results on SUN RGB-D val dataset under different ratios of labeled data.} The best is denoted by \textbf{boldface}, while the second best is \underline{underlined}.} 
\label{table:SUNRGB-D}
\scalebox{0.75}{
\begin{tabular}{c|c|c|c|c|c|c|c|c}
\toprule
\multirow{2}*{Model} & \multicolumn{2}{c|}{1\%} & \multicolumn{2}{c|}{5\%} & \multicolumn{2}{c|}{10\%} & \multicolumn{2}{c}{20\%}  \\
\cline{2-3}\cline{4-5} \cline{6-7} \cline{8-9} 
  & mAP@0.25 & mAP@0.5 & mAP@0.25 & mAP@0.5 & mAP@0.25 & mAP@0.5 & mAP@0.25 & mAP@0.5 \\
\midrule
VoteNet\cite{qi2019deep} & 18.3\(\pm\)1.2 & 4.4\(\pm\)0.4 & 29.9\(\pm\)1.5 & 10.5\(\pm\)0.5 & 38.9\(\pm\)0.8 & 17.2\(\pm\)1.3 & 45.7\(\pm\)0.6 & 22.5 \(\pm\)0.8\\
SESS\cite{zhao2020sess} & 20.1\(\pm\)0.2 & 5.8\(\pm\)0.3 & 34.2\(\pm\)2.0 & 13.1\(\pm\)1.0 & 42.1\(\pm\)1.1 & 20.9\(\pm\)0.3  & 47.1\(\pm\)0.7 & 24.5\(\pm\)1.2 \\
3DIoUMatch\cite{wang20213dioumatch} & 21.9\(\pm\)1.4 & 8.0\(\pm\)1.5 & 39.0\(\pm\)1.9 & 21.1\(\pm\)1.7 & 45.5\(\pm\)1.5 & 28.8\(\pm\)0.7  & 49.7\(\pm\)0.4 & 30.9\(\pm\)0.2 \\
NESIE\cite{wang2023not} & / & / & 41.1\(\pm\)1.2 & 21.8\(\pm\)1.8 & 47.4\(\pm\)0.8 & 29.2\(\pm\)1.2  & \textbf{53.4\(\pm\)0.9} & 31.2\(\pm\)1.3 \\
Diffusion-SS3D\cite{ho2024diffusion} & \underline{30.9\(\pm\)1.0} & \underline{14.7\(\pm\)1.2} & \underline{43.9\(\pm\)0.6} & \underline{24.9\(\pm\)0.3} & \underline{49.1\(\pm\)0.5} & \underline{30.4\(\pm\)0.7} & 51.4\(\pm\)0.8 & \underline{32.4\(\pm\)0.6} \\
\textbf{Ours} & \textbf{32.5\(\pm\)0.8} & \textbf{16.3\(\pm\)1.5} & \textbf{45.7\(\pm\)1.2} & \textbf{26.2\(\pm\)1.1} & \textbf{50.2\(\pm\)0.8} & \textbf{31.7\(\pm\)0.3} & \underline{53.2\(\pm\)1.5} & \textbf{34.0\(\pm\)1.1} \\
\bottomrule
\end{tabular}
}
\end{center}
\end{table}

\subsection{ Comparison with State-of-the-art Methods}
In this section, we compare our Diff3DETR with state-of-the-art approaches on ScanNet dataset and SUN RGB-D dataset.

\begin{figure*}[!htbp]
  \centering
   \includegraphics[width=0.95\linewidth]{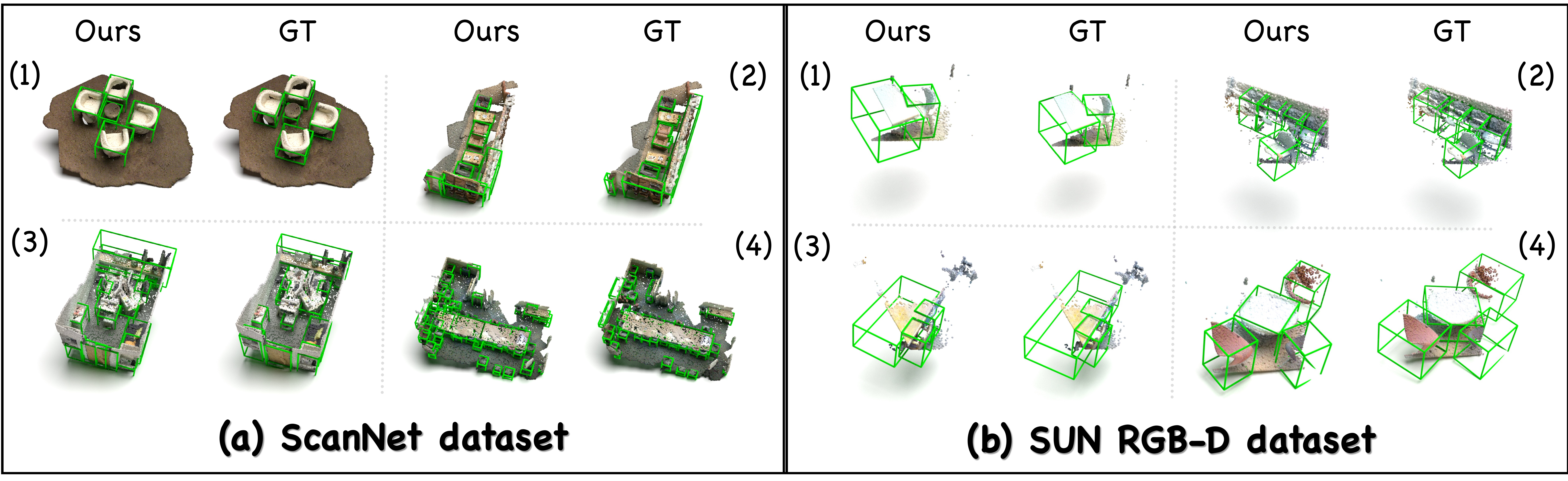}
   \caption{\textbf{The qualitative results on ScanNet and SUN RGB-D datasets.} }
   \label{fig:qualitative}
\end{figure*}

\textbf{Results on ScanNet dataset.}
Table~\ref{table:ScanNet} displays a comparison of our approach against the current state-of-the-art methods on the ScanNet validation dataset, achieving leading results across different ratios of labeled data. Through meticulous design in the generation of object queries and more accurate incremental refinement by the box-aware denoising module for noisy boxes, our method surpasses the best existing method by 1.6\% mAP@0.25 and 1.3\% mAP@0.5 with only 5\% labeled data. Additionally, Figure~\ref{fig:qualitative}(a) presents qualitative results of our method from the ScanNet dataset.

\textbf{Results on SUN RGB-D dataset.}
Table~\ref{table:SUNRGB-D} demonstrates the comparative performance of our method against the current state-of-the-art methods on the SUN RGB-D validation dataset, where it achieves leading results across different ratios of labeled data. Specifically, our method outperforms the best existing method by 1.8\% mAP@0.25 and 1.3\% mAP@0.5 with only 5\% labeled data. Additionally, Figure~\ref{fig:qualitative}(b) presents qualitative results from scenes within the SUN RGB-D dataset, showcasing the capabilities of our method.

\subsection{Ablation Study}
\label{sec:ablationstudy}

\begin{table}[!htbp]
  \begin{center}
    \footnotesize
    \setlength\tabcolsep{5pt}
    \centering
    \caption{\textbf{Evaluation of the model with different designs on ScanNet val dataset.} ``AA'' denotes Agent Adaptor, ``AOQG'' stands for Agent-based Object Query Generator, ``BDM'' refers to Box-aware Denoising Module, ``3DV-PRE'' signifies 3D Vertex Relative Position Encoding, and ``DDIM'' stands for the denoising process.
}
    \begin{tabular}{c|ccccc|c|c}
      \toprule
      & \multirow{2}*{AOQG} &  \multirow{2}*{AA} &   \multirow{2}*{BDM} & \multirow{2}*{3DV-PRE} & \multirow{2}*{DDIM} & \multicolumn{2}{c}{ScanNet (5\%)} \\
      \cline{7-8}
      & & & & & & mAP@0.25 & mAP@0.5 \\
      \midrule
      {[A]} & \ding{55}&\ding{55}&\ding{55}&\ding{55}&\ding{55}&42.2\(\pm\)0.5&27.0\(\pm\)0.3\\
      {[B]} & \ding{51}&\ding{55}&\ding{55}&\ding{55}&\ding{55}&43.1\(\pm\)0.7&27.6\(\pm\)0.5\\
      {[C]} & \ding{51}&\ding{51}&\ding{55}&\ding{55}&\ding{55}&43.5\(\pm\)0.9&27.8\(\pm\)0.6\\
      {[D]} & \ding{51}&\ding{51}&\ding{51}&\ding{55}&\ding{55}&44.3\(\pm\)0.4&28.6\(\pm\)0.3\\
      {[E]} & \ding{51}&\ding{51}&\ding{51}&\ding{51}&\ding{55}&44.6\(\pm\)0.2 &28.8\(\pm\)0.6\\
      {[F]} & \ding{51}&\ding{51}&\ding{51}&\ding{51}&\ding{51}&\textbf{45.1\(\pm\)0.5}&\textbf{29.2\(\pm\)0.5}\\
      \bottomrule
    \end{tabular}
    \label{table:ablation}
  \end{center}
\end{table}

\noindent\textbf{Evaluation of the model with different designs. }
In Table~\ref{table:ablation}, we present a series of ablation studies to validate the effectiveness of our designs. [A] represents the baseline IoU-aware VoteNet model~\cite{wang20213dioumatch} without employing DDIM iterative denoising~\cite{song2020denoising}. [B] illustrates that the agent-based object query generator produces improved object queries which aid the model in achieving a 0.9\% increase in mAP@0.25 and a 0.6\% rise in mAP@0.5. The contrast between [C] and [B] confirms the significance of the agent adaptor for dynamically adapting to the scene. [D] further incorporates the proposed box-aware denoising module, which assists the network in aggregating features from the correct regions and incrementally refining predicted boxes, resulting in an effective increase of 0.8\% mAP@0.25 and 0.8\% mAP@0.5. [E] validates that 3D vertex relative position encoding is conducive to focusing the object queries on point cloud regions surrounding noisy boxes. [F] represents the complete Diff3DETR model which achieves the best performance among all variants.

\begin{figure*}[!b]
  \centering
  \begin{minipage}[b]{0.35\linewidth}
    \centering
    \includegraphics[width=\linewidth]{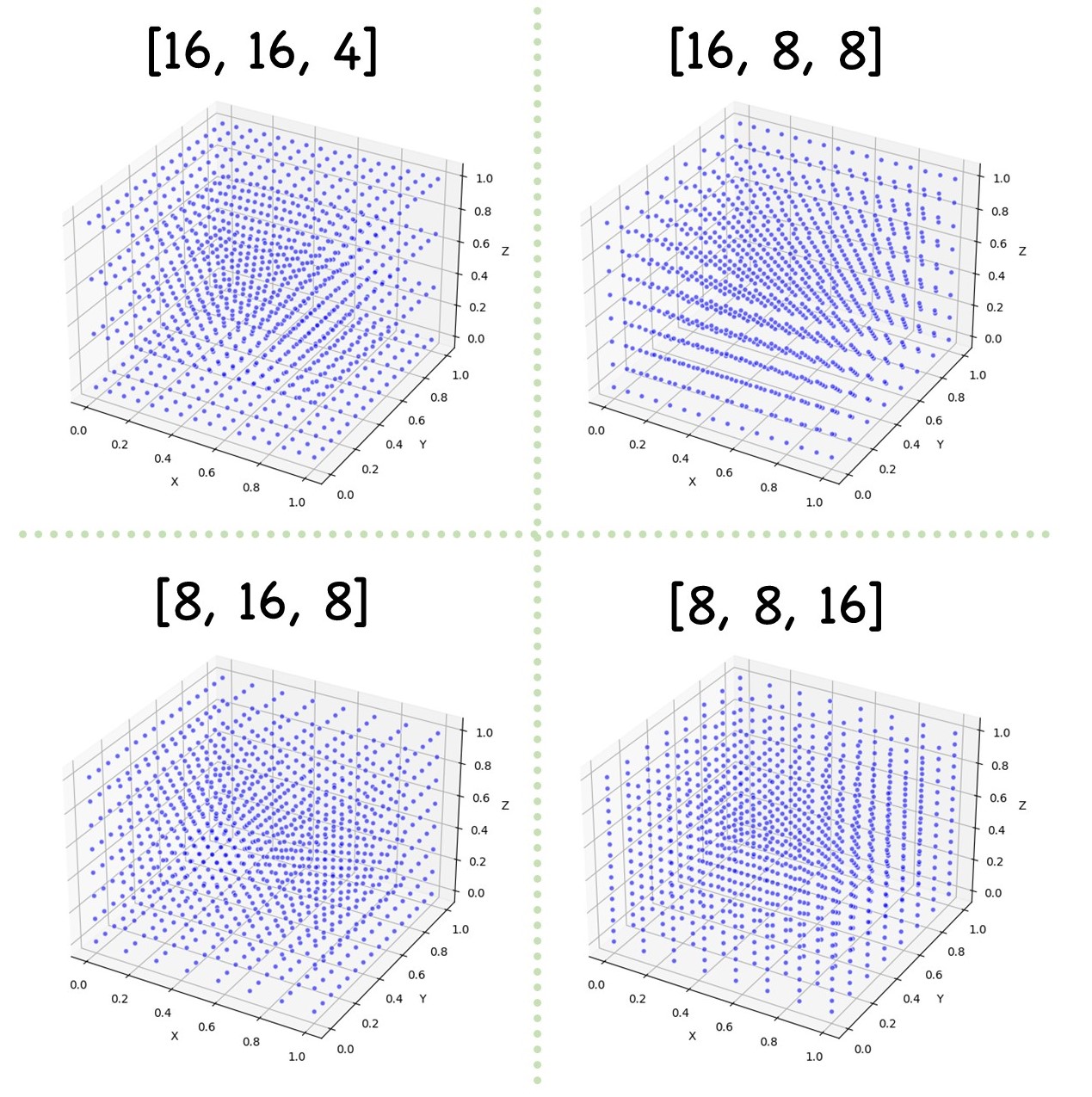}
    \caption{\textbf{Distribution visualization of object agents.}}
    \label{fig:grid}
  \end{minipage}
  \hfill
  \begin{minipage}[b]{0.63\linewidth}
    \centering
    \footnotesize
    \setlength\tabcolsep{5pt}
    \scalebox{0.85}{
      \begin{tabular}{c|cccc}
        \toprule
        Ratio & 5\% & 10\%  & 20\%  & 100\% \\
        \midrule
        \multicolumn{5}{c}{ScanNet mAP@0.25}\\
        \midrule
        $[8, 8, 16]$ & 44.5\(\pm\)0.4 & 50.9\(\pm\)0.9 & 56.2\(\pm\)1.3 & 65.2 \\
        $[8, 16, 8]$ & 44.9\(\pm\)0.3 & 51.2\(\pm\)0.6 & 56.5\(\pm\)0.7 & 65.4 \\
        $[16, 8, 8]$ & 44.8\(\pm\)0.3 & 51.3\(\pm\)1.1 & 56.3\(\pm\)0.9 & 65.4 \\
        $[16, 16, 4]$ & \textbf{45.1\(\pm\)0.5}&\textbf{51.6\(\pm\)1.2}&\textbf{57.0\(\pm\)1.5}&\textbf{65.7}\\
        \midrule
        \multicolumn{5}{c}{ScanNet mAP@0.5}\\
        \midrule
        $[8, 8, 16]$ & 28.5\(\pm\)0.8 & 33.6\(\pm\)0.6 & 37.4\(\pm\)0.2 & 44.1 \\
        $[8, 16, 8]$ & 28.8\(\pm\)0.6 & 33.8\(\pm\)0.9 & 37.9\(\pm\)0.5 & 44.5 \\
        $[16, 8, 8]$ & 28.9\(\pm\)0.3 & 33.8\(\pm\)0.3 & 38.0\(\pm\)0.4 & 44.7 \\
        $[16, 16, 4]$ & \textbf{29.2\(\pm\)0.5}&\textbf{34.2\(\pm\)0.9}&\textbf{38.2\(\pm\)0.7}&\textbf{44.9}\\
        \bottomrule
      \end{tabular}
    }
    \captionof{table}{\textbf{Results under different distributions of object agents.}}
    \label{table:grid}
  \end{minipage}
\end{figure*}

\noindent\textbf{Effectiveness of the agent-based object query generator. }
Object agents are initialized as learnable vectors uniformly distributed in normalized 3D space, and their density across the dimensions of length, width, and height significantly influences the quality of subsequent object query generation. Figure~\ref{fig:grid} illustrates the visualization of grid point distributions under different \([L, W, H]\) resolution settings for length, width, and height. Results from Table~\ref{table:grid} on the ScanNet dataset indicate that the \([16, 16, 4]\) distribution achieves the best performance. This outcome is attributed to the fact that objects within scenes have a notably lower density distribution in vertical height compared to on the horizontal plane, hence allocating more object agents across the horizontal plane aids in modeling a wider variety of spatial semantic information.

\noindent\textbf{Effectiveness of the box-aware denoising module.}
The box-aware denoising module ingeniously integrates the DETR decoder architecture with the DDIM structure, where the number of blocks in both the decoder layer and DDIM significantly affects the model's ability to decode the scene's object boxes. Table~\ref{table:layers} presents ablation studies on the number of blocks in the decoder layer and DDIM, showing that, overall, more decoder blocks and DDIM iterations per cycle yield better detection results. However, increasing the number of blocks in both the decoder layer and DDIM considerably impacts the model's training and inference speed. Therefore, we seek a trade-off between accuracy and computational cost, selecting \#DL=3 and \#DDIM=2 as the final model configuration.

\begin{table}[!h]
    \begin{center}
      \footnotesize
      \setlength\tabcolsep{5pt}
      \centering
      \caption{\textbf{Ablation studies on the number of blocks in the decoder layer and the DDIM.}
      ``\#DL'' and ``\#DDIM'' respectively denote the number of blocks in the decoder layer and the DDIM.
}
      \scalebox{0.75}{
      \begin{tabular}{cc|cccc|cccc}
        \toprule
        \multirow{2}*{\#DL} &\multirow{2}*{\#DDIM} & \multicolumn{4}{c|}{ScanNet mAP@0.25}  & \multicolumn{4}{c}{ScanNet mAP@0.5}\\
        \cline{3-10}
        & & 5\% & 10\% & 20\% & 100\% & 5\% & 10\% & 20\% & 100\%\\
        \midrule
        1 & 1 &42.7\(\pm\)0.2&49.3\(\pm\)1.1&54.9\(\pm\)0.8&63.0&26.5\(\pm\)0.4&31.4\(\pm\)0.8&35.1\(\pm\)0.5&42.1\\
        1 & 2 &43.6\(\pm\)0.6&50.2\(\pm\)1.3&55.6\(\pm\)1.2&64.3&27.7\(\pm\)0.8&32.7\(\pm\)0.8&36.9\(\pm\)0.2&43.8\\
        1 & 4 &43.9\(\pm\)0.7&50.5\(\pm\)0.9&55.9\(\pm\)1.1&64.5&27.7\(\pm\)0.2&32.5\(\pm\)0.7&36.3\(\pm\)0.7&43.7\\
        3 & 1 &44.2\(\pm\)0.3&50.8\(\pm\)0.8&56.2\(\pm\)1.1&64.5&27.9\(\pm\)0.7&33.0\(\pm\)0.7&36.5\(\pm\)0.8&43.5\\
        3 & 2 &45.1\(\pm\)0.5&51.6\(\pm\)1.2&57.0\(\pm\)1.5&65.7&29.2\(\pm\)0.5&34.2\(\pm\)0.9&38.2\(\pm\)0.7&44.9\\
        3 & 4 &45.4\(\pm\)0.4&52.0\(\pm\)1.3&57.3\(\pm\)1.7&66.1&28.8\(\pm\)0.5&34.3\(\pm\)0.7&37.9\(\pm\)0.7&45.3\\
        9 & 1 &45.3\(\pm\)0.9&52.0\(\pm\)0.9&57.0\(\pm\)1.4&65.5&29.0\(\pm\)1.1&33.8\(\pm\)0.8&37.1\(\pm\)1.3&44.9\\
        9 & 2 &\textbf{46.3\(\pm\)0.7}&52.8\(\pm\)1.0&58.2\(\pm\)1.1&66.3&\textbf{30.1\(\pm\)0.6}&35.1\(\pm\)0.6&39.3\(\pm\)0.5&46.2\\
        9 & 4 &46.2\(\pm\)0.5&\textbf{53.4\(\pm\)0.9}&\textbf{58.7\(\pm\)1.2}&\textbf{66.8}&29.7\(\pm\)0.9&\textbf{35.5\(\pm\)0.4}&\textbf{39.4\(\pm\)1.0}&\textbf{46.7}\\
        \bottomrule
      \end{tabular}
      }
      \label{table:layers}
    \end{center}
\end{table}

\subsection{Limitations}
\label{sec:limitation}
The diffusion model diffuses object boxes from ground-truth boxes to a random distribution, and the model learns to reverse this noising process. Applying the diffusion model to semi-supervised 3D object detection tasks offers several inherent merits of the diffusion model. First, the diffusion to a random distribution can generate more diverse pseudo-labels. Second, the denoising process coincides with the decoder process of the detection framework, promoting mutual enhancement. However, the slow denoising process of the diffusion model requires more computational resources for training and inference, hindering the algorithm's potential application in real-time devices and large-scale scenes. A more detailed analysis and discussion regarding model overhead are conducted in the supplementary materials.

\section{Conclusion}
\label{sec:Conclusion}
In this paper, we introduce a novel agent-based diffusion model within a unified DETR framework for semi-supervised 3D object detection. Our proposed Diff3DETR comprises an agent-based object query generator and a box-aware denoising module. The agent-based object query generator is designed to produce object queries that effectively adapt to dynamic scenes while striking a balance between sampling locations and content embedding. Meanwhile, the box-aware denoising module utilizes the DDIM denoising process and the long-range attention in the transformer decoder to incrementally refine bounding boxes, thereby achieving better results. Extensive experiments on the ScanNet and SUN RGB-D benchmarks underline the superiority of our Diff3DETR.

\section*{Acknowledgements}

This work was partially supported by the National Defense Science and Technology Foundation Strengthening Program Funding (Grant 2023-JCJQ-JJ-0219), the National Nature Science Foundation of China (NSFC 62121002), and Youth Innovation Promotion Association CAS.


%
%
\bibliographystyle{splncs04}
\bibliography{egbib}
\end{document}